\documentclass{article}

\usepackage{arxiv}

\usepackage[utf8]{inputenc} 
\usepackage[T1]{fontenc}    
\usepackage{hyperref}       
\usepackage{url}            
\usepackage{booktabs}       
\usepackage{amsfonts}       
\usepackage{nicefrac}       
\usepackage{microtype}      
\usepackage{lipsum}		
\usepackage{graphicx}
\usepackage{natbib}
\usepackage{doi}
\usepackage{cleveref}
\usepackage{subcaption}
\usepackage{adjustbox}
\usepackage{quoting}

\newcommand{\samplemean}{\bar{x}}
\newcommand{\samplesd}{s}
\newcommand{\chatbot}{\emph{Talk2X}}
\newcommand{\aiodbot}{\emph{Talk2AIoD}}
\newcommand{\website}{\emph{Website}}
\newcommand{\studyOne}{Study 1}
\newcommand{\studyTwo}{Study 2}

\title{\chatbot{} - An Open-Source Toolkit Facilitating Deployment of LLM-Powered Chatbots on the Web}

\author{ \href{https://orcid.org/0000-0001-6294-2915}{\includegraphics[scale=0.06]{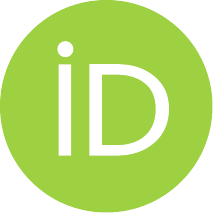}\hspace{1mm}Lars Krupp}\\
	German Research Center for Artificial Intelligence,\\
	RPTU Kaiserslautern-Landau\\
	Kaiserslautern, Germany \\
	\texttt{lars.krupp@dfki.de} \\
	\And
	\href{https://orcid.org/0000-0003-2643-4504}{\includegraphics[scale=0.06]{orcid.pdf}\hspace{1mm}Daniel Geißler} \\
	German Research Center for Artificial Intelligence\\
	Kaiserslautern, Germany \\
    \And
    \href{https://orcid.org/0000-0001-7108-2451}{\includegraphics[scale=0.06]{orcid.pdf}\hspace{1mm}Peter Hevesi}\\
	German Research Center for Artificial Intelligence,\\
	coneno GmbH\\
	Kaiserslautern, Germany \\
    \And
    \href{https://orcid.org/0000-0002-5184-3222}{\includegraphics[scale=0.06]{orcid.pdf}\hspace{1mm}Marco Hirsch}\\
	German Research Center for Artificial Intelligence\\
	Kaiserslautern, Germany \\
    \And
    \href{https://orcid.org/0000-0003-0320-6656}{\includegraphics[scale=0.06]{orcid.pdf}\hspace{1mm}Paul Lukowicz}\\
	German Research Center for Artificial Intelligence,\\
	RPTU Kaiserslautern-Landau\\
	Kaiserslautern, Germany \\
    \And
    \href{https://orcid.org/0000-0002-0698-7470}{\includegraphics[scale=0.06]{orcid.pdf}\hspace{1mm}Jakob Karolus}\\
	German Research Center for Artificial Intelligence,\\
	RPTU Kaiserslautern-Landau\\
	Kaiserslautern, Germany \\
}

\renewcommand{\shorttitle}{\chatbot{}}

\begin{document}
\maketitle

\begin{abstract}
Integrated into websites, LLM-powered chatbots offer alternative means of navigation and information retrieval, leading to a shift in how users access information on the web.
Yet, predominantly closed-sourced solutions limit proliferation among web hosts and suffer from a lack of transparency with regard to implementation details and energy efficiency.
In this work, we propose our openly available agent \chatbot{} leveraging an adapted retrieval-augmented generation approach (RAG) combined with an automatically generated vector database, benefiting energy efficiency. \chatbot's architecture is generalizable to arbitrary websites offering developers a ready to use tool for integration.
Using a mixed-methods approach, we evaluated \chatbot's usability by tasking users to acquire specific assets from an open science repository. \chatbot{} significantly improved task completion time, correctness, and user experience supporting users in quickly pinpointing specific information as compared to standard user-website interaction.
Our findings contribute technical advancements to an ongoing paradigm shift of how we access information on the web.
\end{abstract}


\section{Introduction}
Nowadays, we almost exclusively access information through the internet~\cite{singh2022internet}. Yet, flawed website experiences still frequently plague users~\cite{badwebsites}. As such, with the ever-increasing amount of data available on the web finding relevant information becomes increasingly difficult~\cite{internet_consumption}. Large language models (LLMs) have surfaced as a possible alternative information provider, seemingly enabling convenient access to information. Paired with grounding methods, such as retrieval-augmented generation (RAG)~\cite{lewis2020retrieval}, they become a valid alternative capable of accessing structured information, thus allowing them to cite sources and to reduce problematic LLM-induced hallucinations~\cite{huang2023survey}.

Further pursuing this idea, researchers developed LLM-powered agents~\cite{deng2024mind2web} able to select and populate suited functions 
automatically to solve given tasks. This tool use\footnote{Tool use and function calling are often used interchangeably in the context of LLMs. In this work, we will refer to the action of an LLM calling an external function with inputs and waiting for the results of its execution as function calling.} includes calling functions for data retrieval and analysis, allowing them to cope with complex search queries through self-evaluation of retrieved information.

These technologies steadily integrate into everyday web interaction. As an alternative way of retrieving information from a website, integrated LLM-powered chatbots change how users access information on the web. Yet, lack of transparency regarding implementation details and potentially poor energy efficiency limit the democratization of these technological advances. Especially for information retrieval on websites, large context sizes are necessary to encapsulate the web content, negatively impacting energy efficiency.

In this work, we present an adapted implementation through \chatbot, an openly available function calling agent\footnote{\url{https://github.com/aiondemand/aiod-chatbot}} leveraging an array of optimized search functions that is capable of autonomously executing complex queries on its pre-generated vector database to access relevant information of a website. Leveraging similarity metrics on vector embeddings in its database allows \chatbot{} to work efficiently. Further, its system architecture is generalizable to arbitrary websites offering developers a ready to use tool for integration.


We evaluated the usability of \chatbot{} through a mixed-methods approach in two studies, tasking participants to retrieve assets and information from an open science website. Our first study (N=108) was conducted through Prolific. Choosing a between-subject design, we tasked participants to acquire the information either with the help of \chatbot{} or solely by accessing the website themselves. Our results showed significantly improved task completion times (TCT), effectiveness and usability in favor of \chatbot.
To investigate individual interaction patterns for either conditions (\chatbot{} vs. \website), we conducted a second study (N=12), set up using a think-aloud protocol. 
We found that interaction patterns aligned well with behaviors reported in the literature~\cite{krupp2023challenges}, confirming the usability of our approach. 

Our work contributes (1) technical advancements through an open-source agent\footnotemark[\value{footnote}] to an ongoing paradigm shift of how we access information on the web and (2) presents a systematic usability and sustainability evaluation of its feasibility. 

\section{Related Work}
Offering convenient access through a natural language interface (chatbots), LLMs have become an important source of information for many~\cite{chatgptuse}.
In this section, we elaborate on this paradigm shift, looking at website technology, chatbots and LLMs. 

\subsection{Website Usability}
The design of websites has changed significantly from the first text-based webpages to today's design paradigms~\cite{websites_past}. As the underlying technologies improved~\cite{postel1981internet,huitema1995ipv6} and new technologies were added, such as CSS~\cite{lie1996cascading} or the semantic web~\cite{lassila2001semantic}, the users expectations of a website also underwent changes~\cite{baharum2013users}.

Developers in turn have produced guidelines to ensure consistency for users~\cite{garett2016literature}. This includes how the user interface should look like, the navigation bar~\cite{dos2011usability} and breadcrumbs~\cite{lida2003breadcrumb}.
However, not all websites follow these guidelines, and even when followed, with increasing complexity the difficulty to find something on a website also increases. Leading to users struggling to find information even on well designed websites~\cite{bad_website_design}.
However, recently chatbots are increasingly being used on webpages to support user interaction.

\subsection{Chatbots}
Chatbots have been used in a number of fields, such as education~\cite{kerlyl2006bringing}, medicine~\cite{dharwadkar2018medical} and customer service~\cite{cui2017superagent}. In recent years many companies have started integrating some form of chatbot into their website. Airlines like Lufthansa, who introduced its chatbot Elisa in 2019~\cite{lufthansa_chatbot}, and Air Canada~\cite{aircanada_chatbot} serve as examples. While Elisa is still in use, Air Canada was recently forced to give a customer a discount that its chatbot hallucinated, which ultimately led to the chatbot being taken offline~\cite{aircanada_chatbot2}.
Before the emergence of LLMs, chatbots were rule-based or used natural language processing techniques to try to understand user queries~\cite{adamopoulou2020overview}. In the last years, the amount of work on chatbots has increased drastically with LLMs identified as the next key technology further driving this growth~\cite{lee2023service}. 
However, while the adoption of LLMs allows to build more complex chatbots than ever before, it also opens them up to the weaknesses of these models. Hallucinations, as seen in the airline example, serve as a prominent reminder of their shortcomings.

\subsection{Large Language Models}
Since Vaswani et al.~\cite{vaswani2017attention} published their work on the transformer architecture, large strides have been made in the field. Spurred by the success of ChatGPT, a number of different model lines have been developed, such as LLama~\cite{touvron2023llama,touvron2023llama2,dubey2024llama}, Mistral~\cite{jiang2023mistral,jiang2024mixtral}, and GPT~\cite{openai2024gpt4technicalreport,radford2019language}. More recently, multi-modal LLMs have been gaining traction with models GPT-4V(ision) and PaLM-E~\cite{driess2023palm}.
However, hallucinations are a common issue for these models~\cite{huang2023survey}, leading to the adoption of retrieval-augmented generation (RAG)~\cite{shuster2021retrieval}. In the basic RAG system the user query is first embedded to deploy a semantic similarity search on a vector database, providing the most similar search results to the LLM as context to answer the query. More complex systems, called agents, use tools or call functions during the LLM execution~\cite{masterman2024landscape}.
Some agents use methods like ReAct~\cite{yao2022react}, allowing them to cyclically formulate thoughts, call functions depending on the formulated thoughts and evaluate the results until a stopping condition is reached. Others use linguistic feedback~\cite{shinn2023reflexionlanguageagentsverbal}, chain-of-abstraction reasoning~\cite{gao2024efficienttoolusechainofabstraction} or even have multiple agents cooperate to solve a problem~\cite{shi2024learningusetoolscooperative}. In our work, we adapted some of these techniques, including the general concept of the cyclical ReAct approach in combination with an adapted RAG method. We integrated these ideas into \chatbot{} to increase its efficiency and effectiveness during search.

\section{Methodology}
\label{sec:method}

Our work advocates the introduction of LLM-readable representations of website content. Here, retrieval-augmented generation (RAG) has been established as a way to efficiently provide LLMs with up-to-date sources of information~\cite{lewis2020retrieval}, incidentally also reducing hallucinations~\cite{shuster2021retrieval}. We decided to adapt the RAG-based approach allowing us to store the website content in the form of a vector database, enabling the agent to retrieve information selectively. Our approach eliminates the need to crawl a website repeatedly by storing the website content in the vector database, increasing energy efficiency.

Additionally, we enable the agent to autonomously search for relevant information within its database through a set of functions (\Cref{sec:system}). This reduces the computational overhead (populating the database only once) and decreases the amount of tokens in the context (condensed database representation of the website). Ultimately, our approach allows agents to execute queries more sustainably and at scale.


To confirm the feasibility of this concept, we evaluated our system's usability and user experience in two studies, analyzing and comparing users' interaction behaviors. Our investigation was governed by the following research questions.

\paragraph{RQ1: What are (dis)advantages of using \emph{\chatbot{}} compared to the standard website interaction?}
To answer this question we conducted a quantitative study (N=108) through Prolific, measuring task completion times and usability of \chatbot{} as compared to traditional website interaction (\Cref{sec:studyOne}). This study serves as feasibility evaluation of our concept.

\paragraph{RQ2: How does \emph{\chatbot{}} influence the participants' usage behaviour?}
To answer this research question we conducted a follow-up study (N=12), where we employed a think-aloud protocol (\Cref{sec:studyTwo}) to analyze users' interaction behaviors when using \chatbot. We confirmed whether user behavior with our system differs form reported behavior with LLM-powered chatbots as reported in literature.
\section{Implementation of \chatbot{}}
\label{sec:system}
\chatbot{} is a function calling agent orchestrating a series of efficient search algorithms. It consists of an LLM prompted to plan how to answer a user query, use functions when required, evaluate their results, and repeat this process until a satisfactory output is produced.
\chatbot{} is able to access a vector database containing a website's content (website collection) and, if any, the content of its associated database (asset collection). Using a number of functions, \chatbot{} is able to retrieve the relevant context to answer the question from its database. As such, we only have to crawl the website once and are able to provide the needed context to \chatbot{}.

\begin{figure}[ht]
    \centering
    \includegraphics[width=0.7\linewidth]{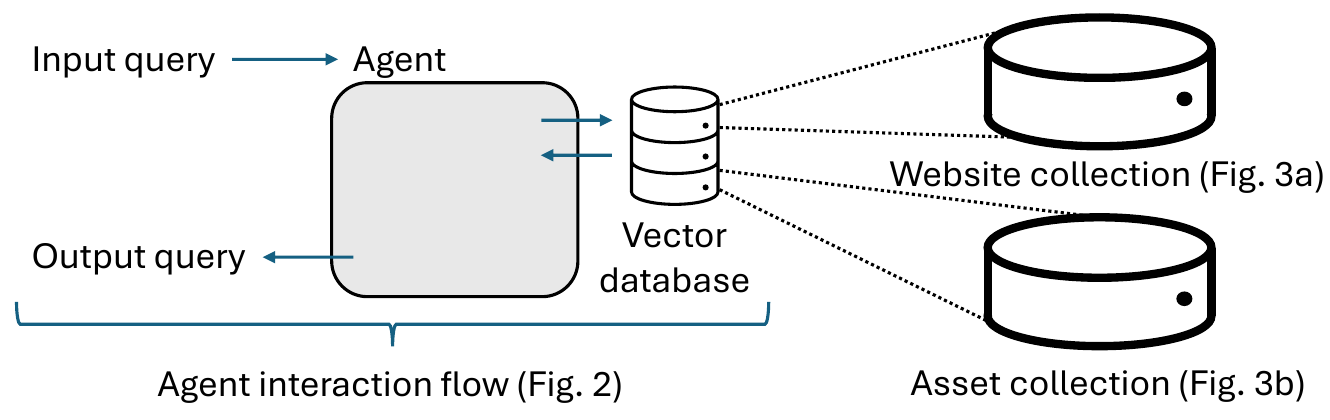}
    \caption{Overview over the interactions and system architecture of \chatbot.}
    \label{fig:complete_flow}
\end{figure}

\chatbot{} consists of three main parts (see \Cref{fig:complete_flow}). The first is the agent interaction flow which models the agent's behavior when interacting with the user. The other two parts are only executed once to create the vector database needed for the agent interaction flow. They build the website collection and the asset collection as depicted in \Cref{fig:complete_flow}.

\begin{figure}[ht]
    \centering
    \includegraphics[width=0.5\linewidth]{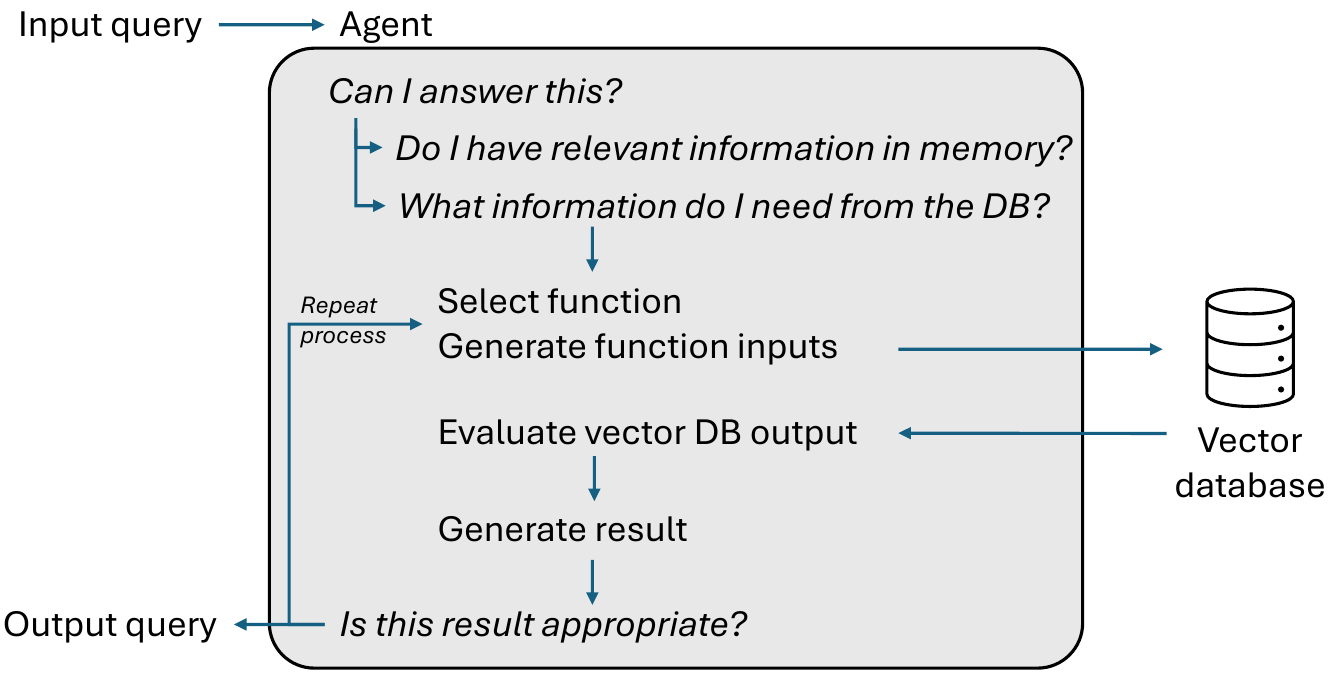}
    \caption{The agents internal interaction flow upon receiving a user query.}
    \label{fig:agent_flow}
\end{figure}

The general idea of \chatbot{} is to build a function calling agent with functions that allow it to retrieve information from the vector database and augment its result generation with it. 
Since the agent interacts with the vector database, instead of just getting information from it once, it is able to send multiple queries or to refine queries that did not yield the necessary information at first (see \Cref{fig:agent_flow}). 

We used LangChain\footnote{\url{https://www.langchain.com/}} to build \chatbot, employing GPT4o-mini\footnote{\url{https://openai.com/index/gpt-4o-mini-advancing-cost-efficient-intelligence/}} as our LLM of choice, due to its performance to cost ratio and speed. The website content was stored in a Chroma\footnote{\url{https://www.trychroma.com/}} database using OpenAI embeddings. The source code is available on github: \url{https://github.com/aiondemand/aiod-chatbot}.

\subsection{Function Calling}
There are multiple ways to make the information contained in the Chroma database available to the agent. We could employ the classical retrieval-augmented generation (RAG)~\cite{shuster2021retrieval} methodology and compute a similarity score of the embedding of each question asked by a user with the elements of both collections giving the LLM the best-fitting information as a context. However, this method lacks flexibility as it always searches both collections exactly once. To improve the agents flexibility and ability to answer complex questions, we use function calling instead allowing the agent to interact with the database multiple times per query.

By using function calling, we allow the LLM to decide based on the user query if it needs to get additional information to answer the question and what kind of information it needs. The LLM is then able to call functions to search for this needed information in a collection and evaluate the result. This approach is more flexible compared to the standard RAG and allows the LLM to search multiple times on the same collection with different parameters until it has retrieved all information needed to answer a user query.

We provide three functions for the LLM to call, giving it the option to search both collections effectively.
\begin{itemize}
    \item \textbf{Website Similarity Search:} Enables the LLM to retrieve relevant information from the website collection using similarity search with L2 as similarity measure.
    \item \textbf{Asset Similarity Search:} Enables the LLM to retrieve relevant information from the asset collection using similarity search with L2 as similarity measure. It can filter by specifying the asset type (publication, dataset, ...) it is looking for.
    \item \textbf{Asset Keyword Search:} Extends the \textbf{Asset Similarity Search} enabling the LLM to specify keyword terms that have to occur within the text returned by the similarity search.
\end{itemize}

\subsection{Database Creation}
\label{sec:impl_db}
\begin{figure}
    \centering
    \begin{subfigure}[t]{0.45\textwidth}
         \centering
         \includegraphics[width=0.8\columnwidth]{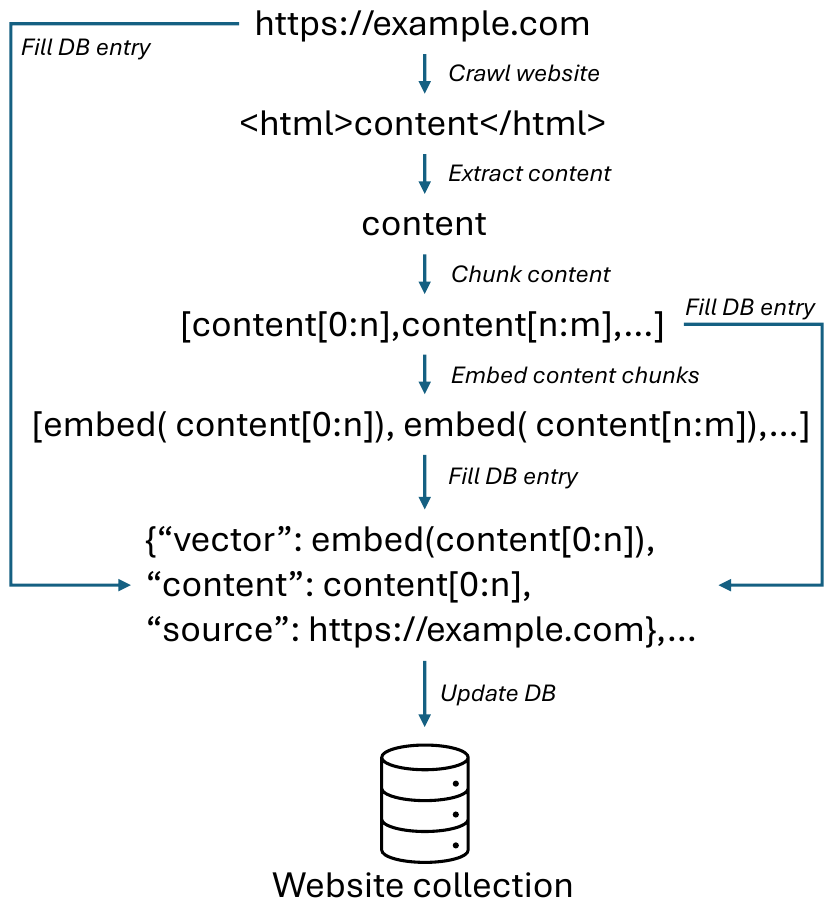}
        \caption{Example process of how a website is crawled including all intermediate steps (website collection). 
        }
        \label{fig:crawl_example}
     \end{subfigure}
     \hfill
    \begin{subfigure}[t]{0.5\textwidth}
         \centering
         \includegraphics[width=\linewidth]{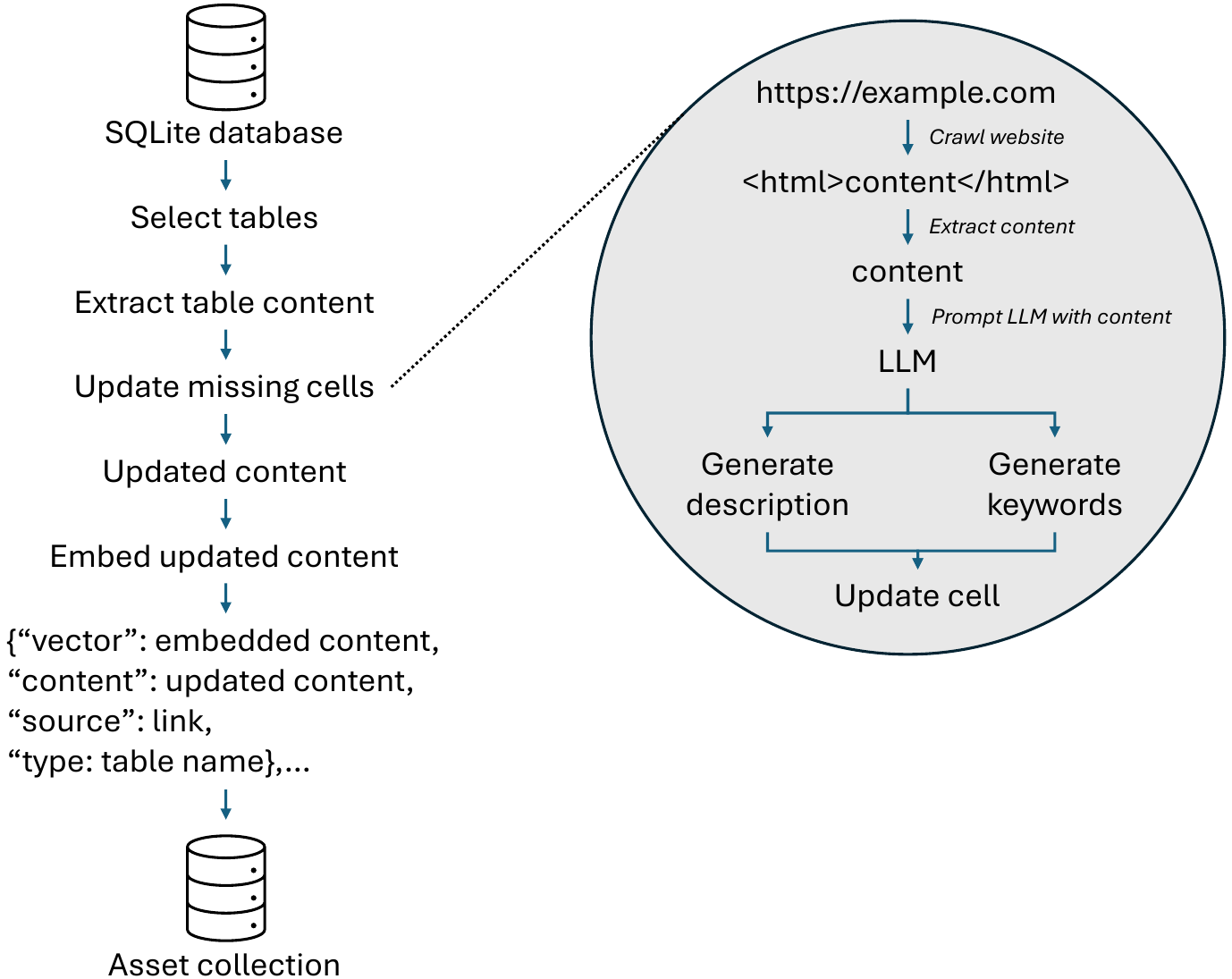}
        \caption{The transformation from our SQLite database to the vector database (asset collection).}
        \label{fig:asset_flow}
     \end{subfigure}
     \caption{Building the website collection (a) and the asset collection (b).}
    \label{fig:website_n_asset}
\end{figure}

The Chroma database used by \chatbot{} consists of two collections. One, called website collection, storing the content of the crawled webpages, where content was extracted from HTML and PDF. This content was then chunked and stored with its associated vector representation and the link to the webpage as source following the process depicted in \Cref{fig:crawl_example}. This automatic process can be used to create a collection for any website.

As lots of websites have one or more associated databases, we additionally present how databases can be integrated into \chatbot. In our evaluation, the website has an associated metadata database (see~\Cref{sec:evaluation}).
The content of this database is stored in the asset collection. We first used the website API to access its metadata database and stored the extracted resources in a temporary SQLite database. We decided to store a subset of the complete metadata database in our SQLite database, restricting ourselves to five tables (\textit{datasets}, \textit{educational resources}, \textit{experiments}, \textit{ml models}, \textit{publications}) with four columns (\textit{name}, \textit{link}, \textit{description}, \textit{keywords}) each. We then checked the description and keywords columns and filled empty cells by crawling the webpage associated with the \textit{link} column and prompting an LLM to write a \textit{description} or a list of \textit{keywords} from the crawled information. For data provenance, we also added a column indicating which cells were automatically generated. Finally, we converted the results into a Chroma collection. This was done by converting \textit{link}, \textit{name}, \textit{keywords} and \textit{description} into one string and calculating its vector representation and storing both in the collection together with the string, as seen in \Cref{fig:asset_flow}. In addition, we stored the resource type (\textit{dataset}, \textit{educational resource}, \textit{experiment}, \textit{ml model} or \textit{publication}) and the source (\textit{link}) as metadata for each row in the collection. Since this process is highly dependent on the structure of the website-associated database there might be a need for adaptations depending on the data store to be transformed into an asset collection.

By providing all collection entries with sources we enabled \chatbot{} to cite its sources of information, making it easier for users to detect hallucinations or get additional information on a topic.

\section{Evaluation}
\label{sec:evaluation}
Through a combination of two user studies, a quantitative study through Prolific (\Cref{sec:studyOne}) and a qualitative think-aloud study in person (\Cref{sec:studyTwo}), we provide a holistic evaluation of our proposed method as implemented through \chatbot.

We chose an open science website (AI on Demand (AIoD): \url{www.aiod.eu}) for the evaluation of \chatbot, calling the resulting chatbot \aiodbot. As the European answer to platforms like Zenodo and GitHub, AI on Demand (AIoD) consists of a multitude of different internal services and webpages, including an associated database, allowing us to highlight the advantages of our concept. This heterogeneous nature makes it an ideal candidate to evaluate \aiodbot.

\subsection{\studyOne: Usability Evaluation Through Prolific}
\label{sec:studyOne}
\studyOne{} was done through Prolific\footnote{\url{https://www.prolific.com/}} employing a between-subject design. Study participants were randomly assigned to either the \aiodbot{} condition or the website condition and tasked with retrieving information and assets from the AIoD website. The participants in the \aiodbot{} condition were allowed to use \aiodbot{} and click on the provided links, while the participants in the website condition were allowed to navigate the website as they wished. Participants were not allowed to use a internet search engine or other external tools. We chose to use Prolific for \studyOne{} due to its high response quality~\cite{Douglas23DataQuality}, ultimately having to discard less then 2\% of all responses.

We tasked the participants with six questions as listed in \Cref{tab:questions_n_tasks}. These questions fall into two categories. First, the meta information questions which require the participants to retrieve knowledge contained within the HTML content of the website. Secondly, the asset questions which require information contained within assets, such as datasets, linked on the website.

\begin{table*}[htb]
	\centering
 \caption{The tasks given to participants. Categorized into meta information questions and asset questions.}
	\label{tab:questions_n_tasks}
	\begin{tabular}{ll}
		\toprule
		\multicolumn{2}{l}{\textbf{Meta information}}\\
		\midrule
		\textbf{Q1} & Who is behind the AIoD platform?\\
		\textbf{Q2} & What is the main purpose of AIoD?\\
		\textbf{Q3} & Name two services/functions that are part of the AIoD platform?\\
        \midrule
        \multicolumn{2}{l}{\textbf{Assets}}\\
        \midrule
        \textbf{Q4} & Please find a dataset that interests you.\\
		\textbf{Q5} & Please try to find the AIoD metadata catalogue.\\
		\textbf{Q6} & Please try to find the mushroom dataset by Jeff Schlimmer.\\
        
		\bottomrule
	\end{tabular}
\end{table*}

\subsubsection{Measurements}
\label{sec:studyOne_measurements}
To evaluate the usability of each condition, we tasked participants to fill in multiple usability questionnaires, depending on their condition. Participants in both conditions were given the UMUX-Lite~\cite{lewis2013umux} questionnaire and the items USAB2 and USAB3 from the perceived website usability questionnaire (PWU)~\cite{FLAVIAN20061} (see \Cref{tab:pwu}). While participants in the website condition where then asked the other five items of the PWU, participants in the \aiodbot{} condition were given five selected questions from the Bot Usability Scale (BUS)~\cite{borsci2022chatbot,borsci2023confirmatory} seen in \Cref{tab:bus_questions}. We decided on this configuration for multiple reasons. Using the questions asked in both conditions, we were able to directly compare them. Note that USAB2 and USAB3 are the only questions also applicable for \aiodbot, hence their choice. Additionally, by including the complete PWU we were able to evaluate the usability of the website in comparison to other existing websites. By adding the questions from the BUS we also received information on how participants viewed the usability of \aiodbot{} in a dedicated chatbot usability questionnaire. Note that administering the whole BUS scale was not possible, as the missing items were not applicable to our scenario.
\begin{table*}[tb]
	\centering
 \caption{Selected Bot Usability Scale (BUS)~\cite{borsci2023confirmatory} questions administered to participants in the \aiodbot{} condition.}
	\label{tab:bus_questions}
	\begin{tabular}{ll}
		\toprule
        \textbf{Abbreviation} & \textbf{Selected Bot Usability Scale (BUS) questions}\\
		\midrule
		\textbf{BUS3} & Communicating with the chatbot was clear.\\
		\textbf{BUS6} & I find that the chatbot understands what I want and helps me achieve my goal.\\
		\textbf{BUS8} & The chatbot only gives me the information I need.\\
        \textbf{BUS9} & I feel like the chatbot’s responses were accurate.\\
		\textbf{BUS11} & My waiting time for a response from the chatbot was short.\\
        
		\bottomrule
	\end{tabular}
\end{table*}

\begin{table*}[tb]
	\centering
 \caption{Question items in the perceived website usability (PWU) questionnaire~\cite{FLAVIAN20061}. Note that USAB2 and USAB3 were also administered for the\aiodbot{} condition.}
	\label{tab:pwu}
    \begin{adjustbox}{max width=\textwidth}
	\begin{tabular}{lcl}
		\toprule
        \textbf{Abbreviation} & \textbf{\aiodbot{}} & \textbf{Question}\\
        \midrule
		\textbf{USAB1} & &In this website everything is easy to understand.\\
        \textbf{USAB2} &\checkmark &This website/chatbot is simple to use, even when using it for the first time.\\
        \textbf{USAB3} &\checkmark &It is easy to find the information I need from this website/with this chatbot.\\
        \textbf{USAB4} & &The structure and contents of this website are easy to understand.\\
        \textbf{USAB5} & &It is easy to move within this website.\\
        \textbf{USAB6} & &The organisation of the contents of this site makes it easy for me to know where I am when navigating it.\\
        \textbf{USAB7} & &When I am navigating this site, I feel that I am in control of what I can do.\\
		\bottomrule
	\end{tabular}
    \end{adjustbox}
\end{table*}

\subsubsection{Apparatus}
\label{sec:apparatus}
\begin{figure}[ht]
    \centering
    \includegraphics[width=0.5\linewidth]{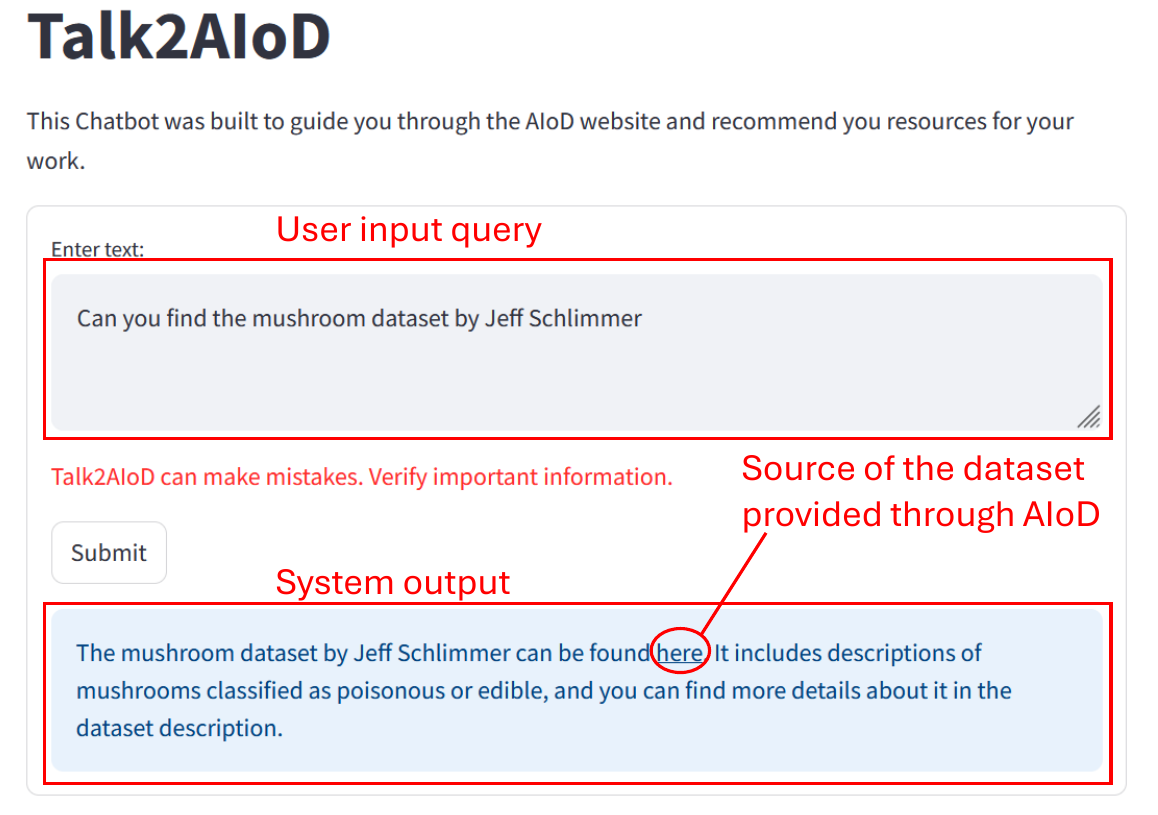}
    \caption{The user interface of \aiodbot{} depicting a search mask and response field.}
    \label{fig:screenshots}
\end{figure}

For both our user studies, we set up a server with CASE~\cite{case24}, our survey tool of choice. We deployed \aiodbot{} on a Gunicorn\footnote{\url{https://gunicorn.org/}} server. The frontend was implemented using streamlit\footnote{\url{https://streamlit.io/}} and is depicted in \Cref{fig:screenshots}. 
In addition we ran a ChromaDB CLI as vector database (\Cref{sec:impl_db}) that was connected to \aiodbot. We logged all interactions with \aiodbot, with timestamps and user id, allowing for a seamless and complete understanding of the participants and agents behavior. We also implemented a simple memory for the chatbot, allowing it to answer questions about previously produced results and improving its conversational abilities.

\subsubsection{Procedure}
\begin{figure}
    \centering
    \includegraphics[width=\linewidth]{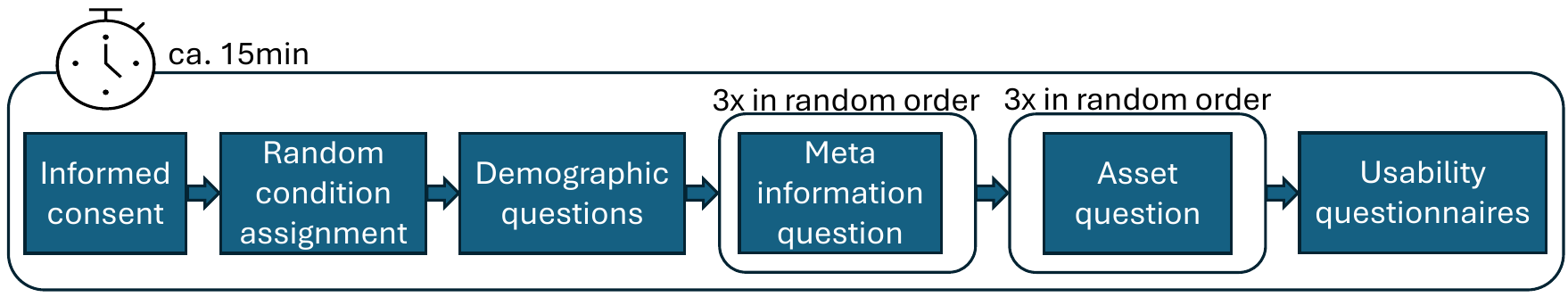}
    \caption{The procedure of \studyOne. The meta information and asset questions can be found in \Cref{tab:questions_n_tasks}.}
    \label{fig:quantitative_procedure}
\end{figure}

The whole study procedure is visualized in \Cref{fig:quantitative_procedure} and took approximately 15 minutes to complete. Participants were first asked demographic questions (age, gender, country of residence, country of origin) followed by their familiarity with the AI on Demand (AIoD) platform. They were randomly assigned to either the \aiodbot{} or the website condition. Then, participants were provided with the three questions asking for meta information, followed by the three asset acquisition tasks, as seen in \Cref{tab:questions_n_tasks} both in randomized order. Afterwards, all participants were asked to answer their condition-specific usability questionnaires. The study procedure was approved by the DFKI ethics board.


\subsubsection{Participants}
We collected an initial 110 responses from participants. After removing duplicates, 108 responses were used for analysis. Participants in the \aiodbot{} condition (N=57; Age: $\samplemean=31.8\,y$, $\samplesd=10.7\,y$; 24 female, 33 male) reported a low level of experience\footnote{Measured on a visual analog scale from 0 to 100.} with AIoD ($\samplemean=17.1$, $\samplesd=17.9$) similar to the website condition (N=51, Age: $\samplemean=33.0\,y$, $\samplesd=9.50\,y$,30 female, 21 male), who also report a low level of AIoD experience ($\samplemean=17.9$, $\samplesd=23.8$). The unfamiliarity with AIoD is beneficial for our evaluation eliminating biases. People from more then 20 different countries and four different continents participated in this study. Of these, 64 were born in Europe, 12 in North America, 22 in Africa and 10 in Asia. The participants reported their residence in 16 countries and three different continents. 91 participants resided in Europe, 16 in North America and one in Asia. All participants were reimbursed with an equivalent of 9\pounds{} per hour.

\subsubsection{Results}
We analyzed task completion times, response correctness and usability as measured by the administered questionnaires using one-way ANOVAs after ART~\cite{wobbrock2011aligned} to rank-align the data if normality was violated.

\begin{figure}[ht]
    \centering
    \includegraphics[width=\linewidth]{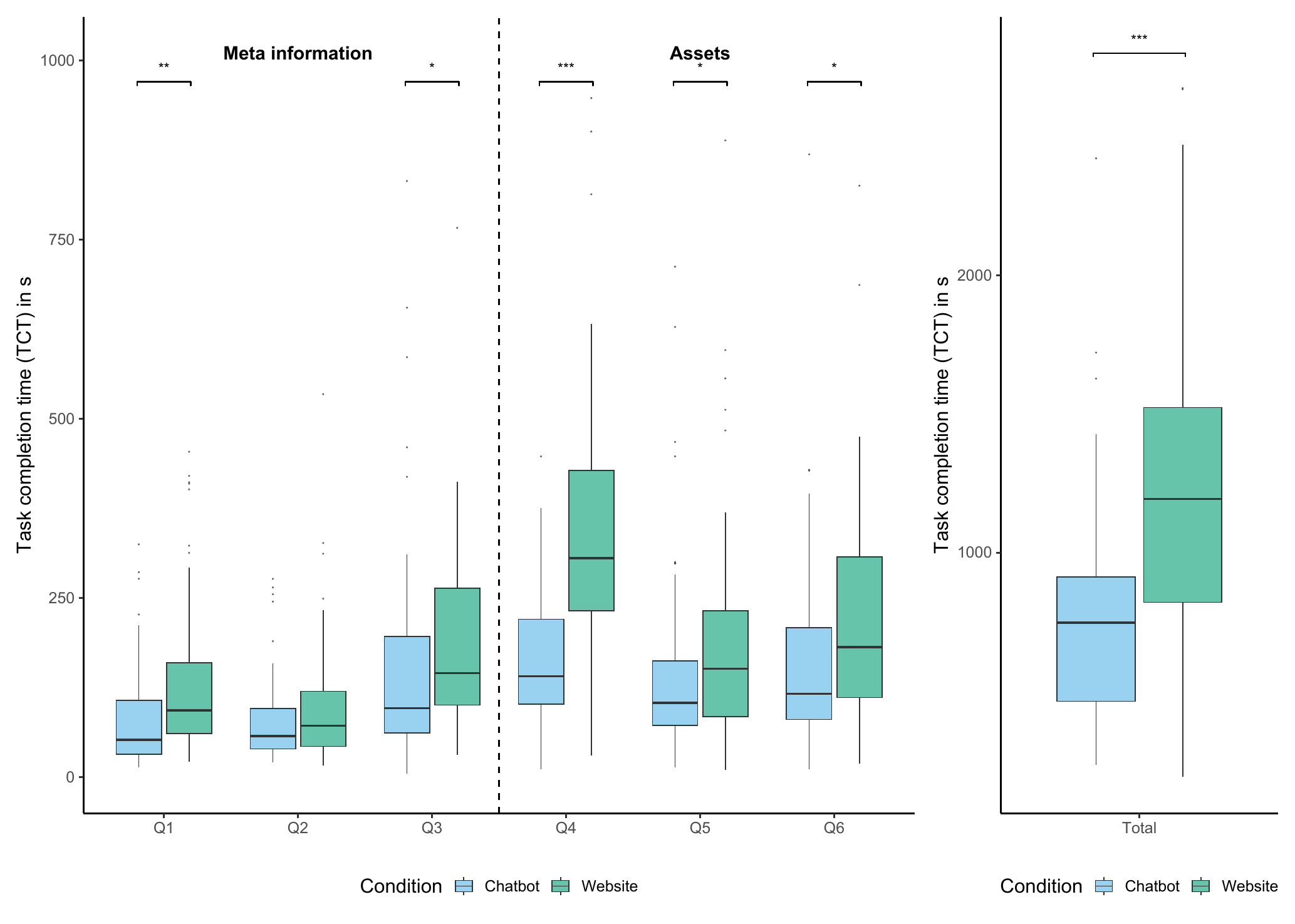}
    \caption{Boxplot depicting the TCT for each question with * indicating significance levels. Questions can be found in \Cref{tab:questions_n_tasks}. Total is the sum of TCT per participant.}
    \label{fig:tct_per_question}
\end{figure}

\paragraph{Task completion time (TCT)} We found a statistically significant difference between the conditions ($F(1,106)=25.87$, $p<1e^{-5}$) with participants in the chatbot condition ($\samplemean=790\,s$, $\samplesd=411\,s$) completing the survey significantly faster compared to participants in the website condition ($\samplemean=1233\,s$, $\samplesd=544\,s$). In addition to this, we analyzed the data for each question individually. We found statistically significant differences in five of six questions, where the average TCT is lower for participants in the \aiodbot{} condition. A visualization of these results is provided in \Cref{fig:tct_per_question}. 


\begin{figure}[ht]
    \centering
    \includegraphics[width=0.4\linewidth]{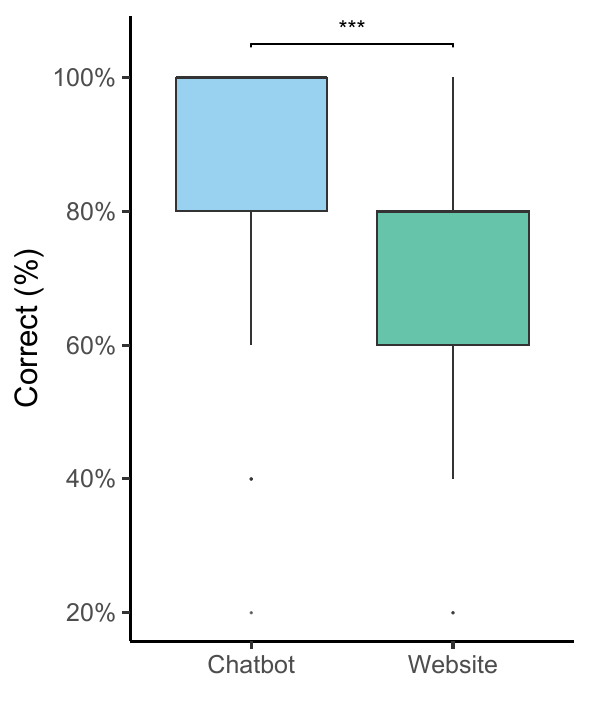}
    \caption{Boxplot depicting the amount of correct participant responses in percent per condition.}
    \label{fig:correctness_boxplot}
\end{figure}

\label{sec:response_correctness}
\paragraph{Response Correctness} While participants using the website produced correct answers in 69\% of cases, participants using \aiodbot{} answered correctly 88\% of the time, as seen in \Cref{fig:correctness_boxplot}. 
We found significant differences for response correctness given the condition ($F(1,106)=35.234$, $p<1e{-7}$). Note that we excluded Q4 ("Please find a dataset that interests you.") from this analysis, since it is an exploratory question.

\begin{figure}[ht]
    \centering
    \includegraphics[width=.8\linewidth]{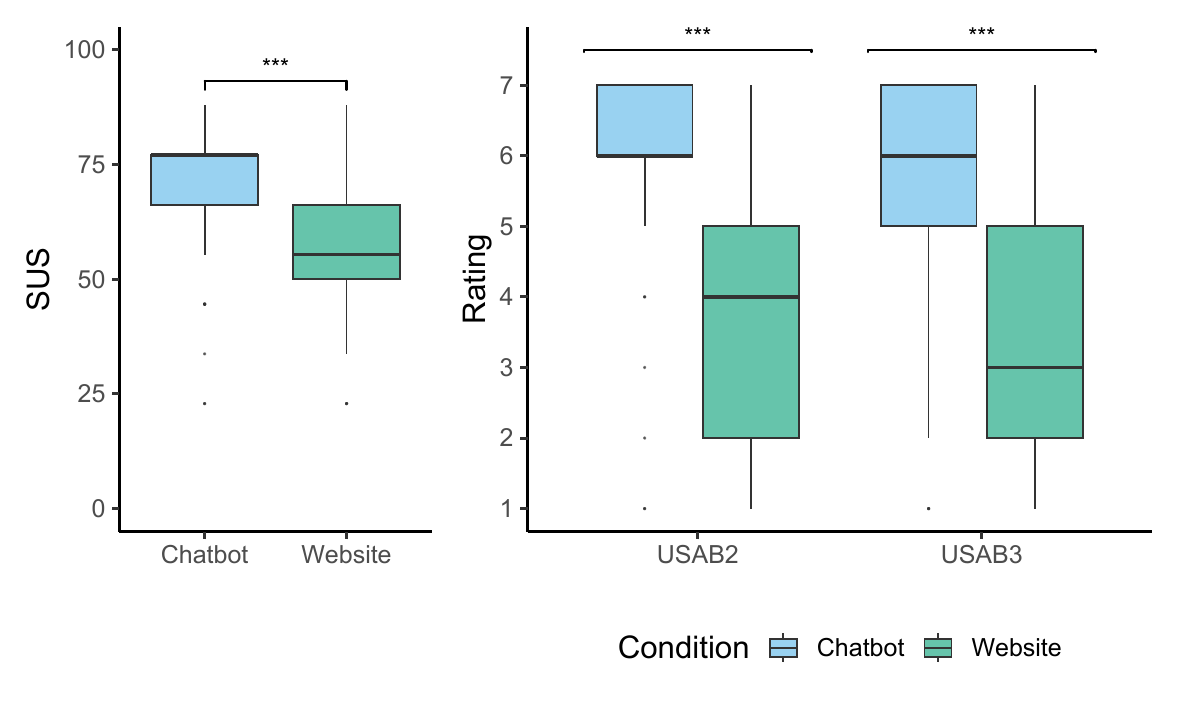}
    \caption{SUS scores (left) and PWU questions (right) given condition. Significance levels marked with *.}
    \label{fig:sus_n_pwu}
\end{figure}

\label{sec:usability}
\paragraph{Usability} To evaluate the \textsc{usability} of the conditions we started by transforming the UMUX-Lite questionnaire into SUS~\cite{lewis2015investigating}.
We found a significant difference between the conditions ($F(1,106)=18.845$, $p<1e{-3}$). The difference between the chatbot condition ($\samplemean=70.8$, $\samplesd=16.4$) and the
website condition ($\samplemean=58.8$, $\samplesd=16.8$) is visualized in \Cref{fig:sus_n_pwu}.



\begin{figure}[ht]
    \centering
    \includegraphics[width=0.4\linewidth]{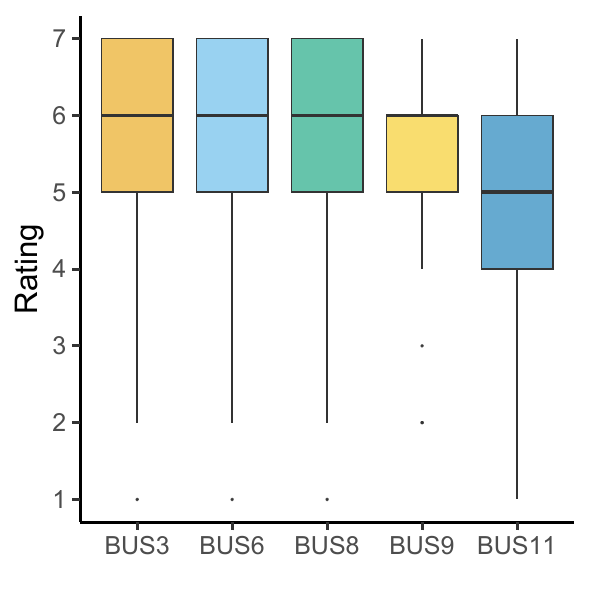}
    \caption{Boxplot depicting the results of the five BUS questions.}
    \label{fig:bus_mean_sd}
\end{figure}

We administered the PWU questionnaire~\cite{FLAVIAN20061} using a seven point Likert-scale. The result ($\samplemean=3.87$, $\samplesd=1.51$) indicates an average website usability~\cite{thielschtoolbox}. The \aiodbot{} usability was evaluated using five selected questions from the Bot usability scale (BUS-11) questionnaire~\cite{borsci2022chatbot,borsci2023confirmatory}, as seen in \Cref{tab:bus_questions}.
We decided to select these questions since a part of the BUS-11 scale deals with comparing chatbots between each other, which is not applicable in our case. The questions were rated on a seven point Likert-scale and showed above average values for all questions (see \Cref{fig:bus_mean_sd}).

Finally, we evaluated the two items from the PWU that were administered in both conditions, USAB2 and USAB3 (\Cref{tab:pwu}). We measured a significant difference between our conditions for both USAB2 ($F(1,106)=59.2$, $p<1e{-3}$) and USAB3 ($F(1,106)=35.58$, $p<1e{-3}$). Participants in the chatbot condition reported higher values for both USAB2 ($\samplemean=5.91$, $\samplesd=1.43$) and USAB3 ($\samplemean=5.42$, $\samplesd=1.69$) compared to the website condition ($\samplemean_{USAB2}=3.67$, $\samplesd_{USAB2}=1.76$, $\samplemean_{USAB3}=3.45$, $\samplesd_{USAB3}=1.83$) as can be seen in \Cref{fig:sus_n_pwu}.



\subsection{\studyTwo: Investigating Interaction Patterns}
\label{sec:studyTwo}


\studyTwo{} was conducted in person as a think-aloud study~\cite{blandford2016qualitative} using a between-subject design. For this qualitative study, all participants had a background in research, a target group of the AIoD platform, allowing us to collect expert feedback. We collected data using screen and audio recordings, and the \aiodbot{} logs.

After the study was conducted, we transcribed the audio files using Whisper~\cite{radford2022robustspeechrecognitionlargescale}. We conducted a focused analysis based on the pragmatic approach by Blandford et al.~\cite{blandford2016qualitative}. Two researchers coded an initial representative set of transcripts and compared their results. Differences were discussed and an agreement for an initial coding tree was reached. One researcher then coded the remaining transcripts using the coding tree. A final discussion consolidated the coding tree.

\subsubsection{Procedure}
We used the same apparatus as in \studyOne{} but adapted the study protocol. We removed the usability questionnaires, country of origin and country of residence. Instead we asked for the field of research in the demographics and gave the participants five minutes to get acquainted with their condition before asking questions. We later asked for their impressions about the system they used. The study took approximately 20 minutes and was approved by the DFKI ethics board. 

\begin{figure}[ht]
    \centering
    \includegraphics[width=\linewidth]{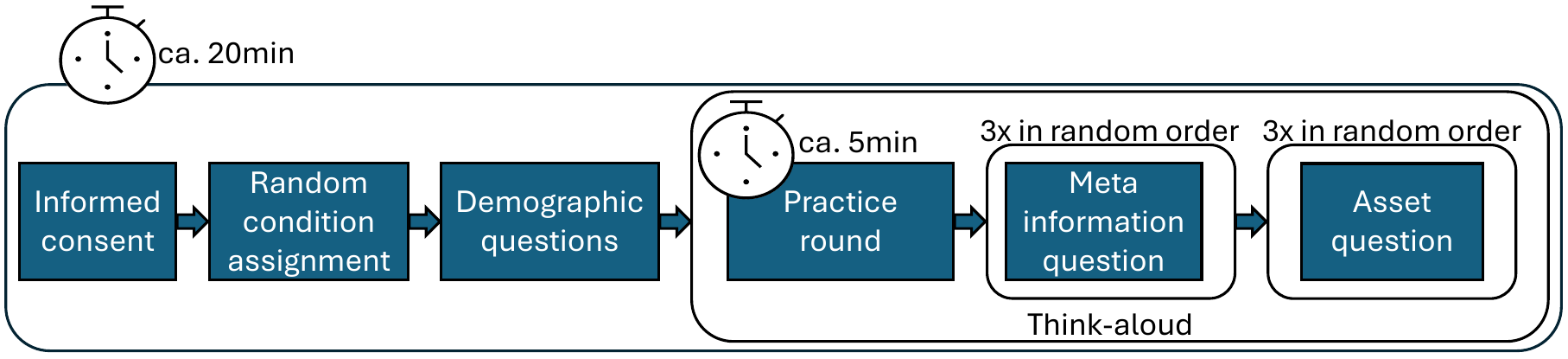}
    \caption{The procedure of \studyTwo. The meta information and asset questions can be found in \Cref{tab:questions_n_tasks}.}
    \label{fig:qualitative_procedure}
\end{figure}

\subsubsection{Participants}
A total of N=12 participants took part in \studyTwo.
Of the participants in the \aiodbot{} condition (N=7; Age: $\samplemean=29.9\,y$, $\samplesd=4.06\,y$; 2 female, 5 male) and \website{} condition (N=5; Age: $\samplemean=31\,y$, $\samplesd=4.80\,y$; 0 female, 5 male), only one had heard of AIoD at all.
The fields of research reported by participants were computer science (4), embedded intelligence (3), physics and computer science (2), artificial intelligence (2) and human computer interaction (1).

\subsubsection{Results}
Our analysis surfaced two main themes: \textsc{Interaction Patterns} and \textsc{Usability Issues}. We elaborate on the content of each theme and associated quotes.

\paragraph{\textsc{Interaction Patterns}}
We identified several unique \textsc{Interaction Patterns} for the different conditions. In part, they are directed by the given interfaces. As such, the contrasting patterns depict specific methods on how users leverage these system.
For our \website{} condition, users explored the website through its interface elements, mostly focusing on built-in search functions, the navigation bar as well as relying on web standards.
\begin{quoting}
"(...) from the left side, it's very clear. Yeah. I can, for example, I'm not sure the metadata set is one of the very important topic of the website. If it is, I should find it from the menu."~(P4)
\end{quoting}
\begin{quoting}
"If you were on the right pages you had the ability to search for something like the data sets." (P7)
\end{quoting}
These interaction patterns align with website guidelines~\cite{garett2016literature,baharum2013users} allowing for generalized interaction paradigms across websites. However, when not applicable, participants reported on fall back solutions, often degenerating in a random exploration of website content.
\begin{quoting}
"I went to the main page. Scrolled down. What was that? Marketplace. Found. Found marketplace. Went back because I think marketplace was the wrong place. Sure. Then I found. Something related. On the main page. And searched. Next. Pages." (P7)
\end{quoting}
In our \aiodbot{} condition, user behavior closely followed interaction paradigms from the literature~\cite{krupp2024unreflected,arora2024analyzing}. Nearly all users simply directed their question at \aiodbot, receiving a concrete and correct answer.
\begin{quoting}
 "Give me a medical dataset. And it worked. Good. I think that's correct." (P2)
\end{quoting}
However, users also were incited to probe and challenge the chatbot, testing out its limitations, a behavior well represented in related work~\cite{krupp2023challenges}.

\paragraph{\textsc{Usability Issues}}
Our interviews also surfaced a few \textsc{Usability Issues} that users struggled with during their tasks.
For our \website{} condition in particular, users complained about navigation issues and often experienced expectation mismatches regarding their own ideas on where to find specific information.
\begin{quoting}
"But the problem is, OK, so according to the logic, so I go into the metadata. Maybe there's some menu. I come back to the metadata set, metadata information. So actually, I didn't find it from here. So maybe it should have some information about such metadata" (P4)
\end{quoting}
Expectation mismatches also occurred in our \aiodbot{} condition, but of a different nature. Here, users were confused that subtle changes to their prompt did not yield different results from \aiodbot.
 \begin{quoting}
 "So I ask it simply again, the same thing because the website looked a little weird to me and provides me with the same thing." (P2)
 \end{quoting}
Interestingly, this caveat was also used to positive effect in a strategy that allowed for verification of initial results.
 \begin{quoting}
 "Let's ask some additional questions. What groups are part of... Behind the AI on demand platform. Group creating AI on demand platform. What groups are part of... What groups are creating... Yes. And that's the same answer. I think unite AI community and promote European values. Alright." (P2)
 \end{quoting}
Participants additionally reported slow responses from \aiodbot{} that hindered their process.

\section{Discussion}
We discuss our findings focusing on the comparative performance of \chatbot{} and simple website search, and the limitations of our work. We further provide deployment scenarios for \chatbot.

\subsection{\chatbot{} Outperforms Traditional Website Search}
Our results of \studyOne{} (see Section \ref{sec:studyOne}) showed that agents like \chatbot{} can significantly reduce the time it takes to acquire information for users, increasing effectiveness and usability compared to website search (\textbf{RQ1}). This improvement mainly manifests in the employed interaction strategies (see \studyTwo, Section \ref{sec:studyTwo}). Participants with the \aiodbot{} condition generally asked \textit{direct questions} (\textbf{RQ2}), an expected behavior previously reported in literature~\cite{krupp2024unreflected}. While negatively connotated there, this behavior can be interpreted positively in our case. Participants \textbf{explicitly search for specific information at little expense} and additionally get the information sources enabling them to validate the agents response, cf.~\cite{wu2024llmsciterelevantmedical}.

The extend of this adapted behavior is visible in \Cref{fig:tct_per_question}. \textbf{Participants using \aiodbot{} completed their tasks significantly faster}, owed to the direct method of accessing the website's content provided by \chatbot{} (\textbf{RQ1}). Instead of sifting through the information themselves participants using \aiodbot{} were essentially able to outsource this process, saving time.
Additionally, we found that using \aiodbot{} increased the amount of correct responses (\textbf{RQ1}).

In addition \textbf{\aiodbot{} was considered significantly more usable} in comparison to the website (see \Cref{fig:sus_n_pwu}). On first glance, the reason for this could be that the website usability was lacking, however, we were able to assess an average PWU score for the website. As such there is a real user preference for \chatbot{} (\textbf{RQ1}). This conclusion is also supported by the high values achieved by \aiodbot{} in the BUS items we evaluated. 
In particular participants reported that \aiodbot{} understood their questions (BUS6), was accurate (BUS9) and only gave them the information they needed (BUS8) (\textbf{RQ1}).

Agents have the distinct advantage of being able to utilize a user's conversation history. As such, \chatbot{} is also able to converse with its user narrowing down requests and finding assets related to the conversation topic. This advantage is present in \aiodbot{} for the task completion time of Q4 (see \Cref{fig:tct_per_question}) in which participants were asked to find a dataset that interests them. This question gave participants the freedom of choice and encouraged them to explore the available datasets. We found a strong statistical significant difference between the conditions for Q4 with participants using \aiodbot{} finishing the task in about half the time (\textbf{RQ1}). 
While conversation flows have been shown to be beneficial in multiple domains, such as surveys~\cite{kim2019comparing} and customer service~\cite{ngai2021intelligent}, the design qualities of this approach remain to be investigated in future research, in particular for asset-heavy website like open science repositories. With increasing LLM capability the advantages of web agents over traditional website use will likely increase, especially in complex information aggregation tasks.

\subsection{Limitations}
In our evaluation, we implemented \chatbot{} for only one website (\aiodbot) consisting of 19 webpages and its associated resource database. We have chosen \url{aiod.eu} as an example of a typical open science repository. Yet, our approach is not limited to any particular website (see~\Cref{sec:system}) and can be easily adapted to other websites. Although, parts of the pipelines, in particular the asset collection (see \Cref{fig:asset_flow}), potentially require manual changes according to how data is stored or accessed on the website.

Likewise, we only used a limited set of questions to evaluate our approach. While we selected a variety of different questions, there is an theoretically unlimited amount of possible questions that users may ask. Our questions were informed through actual \url{aiod.eu} users and explicitly tested both the website collection and the asset collection, making sure that integral components of \chatbot{} were evaluated (see~\Cref{sec:system}). In addition, to counteract restrictions through specific questions, we chose to include Q4 (\Cref{tab:questions_n_tasks}), an exploratory question tasking user to find an interesting dataset of their liking.

Finally, while RAG~\cite{shuster2021retrieval} increases the reliability of \chatbot, hallucinations can still occur. To further increase the reliability and auditability of \chatbot{}, it provided the links to all sources used for its responses. This allowed users to verify the response and check if a hallucination occurred, cf.~\cite{wu2024llmsciterelevantmedical}.

\subsection{Deploying and Using an LLM-Powered Chatbot - \chatbot{} Applications}
In our work we have shown that agents improve the user experience on websites, leading to increased acceptance and widespread use of the technology. To further illustrate this aspect, we present two use case scenarios involving \chatbot{} from two different perspectives: developer and user.

\subsubsection{Deploying a \textit{Talk2X} Agent}
Christian has been hosting a blog about gardening for several years now. He is quite proud of his achievement as it is one of the most extensive on the internet, consisting of hundreds of detailed entries. However, users have contacted him, concerned with the increasing difficulty to find information about specific plants or gardening techniques on his blog. To alleviate this issue, Christian decides to implement a version of \chatbot{} tailored for his blog. He builds the vector database based on his blog entries and integrates the user interface into his website. His users have greatly appreciated the added simplicity in finding, accessing, and aggregating information within his blog.

\subsubsection{Interacting with \chatbot}
Ada is a computer scientist with a focus on human activity recognition. She has been searching for datasets with specific properties on the internet for a while and stumbled across the AIoD website. Ada notices the chatbot \aiodbot{} to interact with. Curious, she asks for the specific characteristics of a dataset she has been searching for. The chatbot thinks for a moment before it provides a few links to datasets available through the digital library associated with AIoD. One of the links provides exactly the information Ada has been looking for. She is satisfied about the speed of the process and visits the page more often in the future in search for suitable datasets.

\section{Conclusion}

We present \chatbot, a function calling agent that aids users in quickly pinpointing required information on modern-day websites. \chatbot's architecture is designed for efficiency and generalizes to arbitrary websites. We evaluated our system in two studies showing a user preference for \chatbot{}. Participants using \chatbot{} were faster and delivered more correct results than their counterparts executing a baseline website search. Our openly available agent serves as a tool for web hosts to integrate LLM-powered chatbots.




\bibliographystyle{unsrtnat}
\bibliography{bibliography}
\end{document}